\documentclass[11pt]{article}
\usepackage[preprint]{acl}
\usepackage{times}
\usepackage{latexsym}
\usepackage[T1]{fontenc}
\usepackage[utf8]{inputenc}
\usepackage{microtype}
\usepackage{inconsolata}
\usepackage{graphicx}
\usepackage{amsmath}
\usepackage{amssymb}
\usepackage{booktabs}
\usepackage{tabularx}

\usepackage{url}
\usepackage{xcolor}
\usepackage{colortbl}
\usepackage{multirow}
\usepackage{array}
\usepackage{ragged2e}
\usepackage{dblfloatfix}
\usepackage{float}
\usepackage{placeins}

\newcolumntype{Y}{>{\RaggedRight\arraybackslash}X}

\newcommand\blfootnote[1]{%
  \begingroup
  \renewcommand\thefootnote{}\footnote{#1}%
  \addtocounter{footnote}{-1}%
  \endgroup
}

\title{Specialization, Diversity, and Scaling in Child-Scale Language Models}

\author{Steven Y. Feng, Michael C. Frank \\
Stanford University\\
\texttt{\{syfeng,mcfrank\}@stanford.edu} \\
}

\begin{document}
\maketitle
\begin{abstract}
Modern language models are trained on far more language than children experience, but we know much less about learning from small, naturalistic datasets. We train small GPT-2 models on transcripts of speech with children from dense home recordings (BabyView) and on synthetic child-directed dialogue (TinyDialogues), varying the amount, source, and diversity of training data. Models trained on one child's language environment predict that environment substantially better than other children's environments. Much of this specialization is lexical and distributional: a simple 5-gram model reproduces the cross-child pattern ($r=.91$). Training on data from multiple children clearly improves prediction of pooled held-out language relative to training on a single child's input without a similarly clear improvement on grammatical benchmarks. Across data sizes, naturalistic and synthetic training inputs also produce different grammatical and lexical-semantic patterns, with important differences between grammatical phenomena. Together, these results show that small-data language learning depends not only on how much language a model sees, but on which environments it comes from and how well that input matches what the model is asked to predict.
\blfootnote{
Code and data: \href{https://github.com/styfeng/child-scale-lm}{github.com/styfeng/child-scale-lm}}
\end{abstract}

\begin{figure*}[t]
\centering
\includegraphics[width=.98\textwidth]{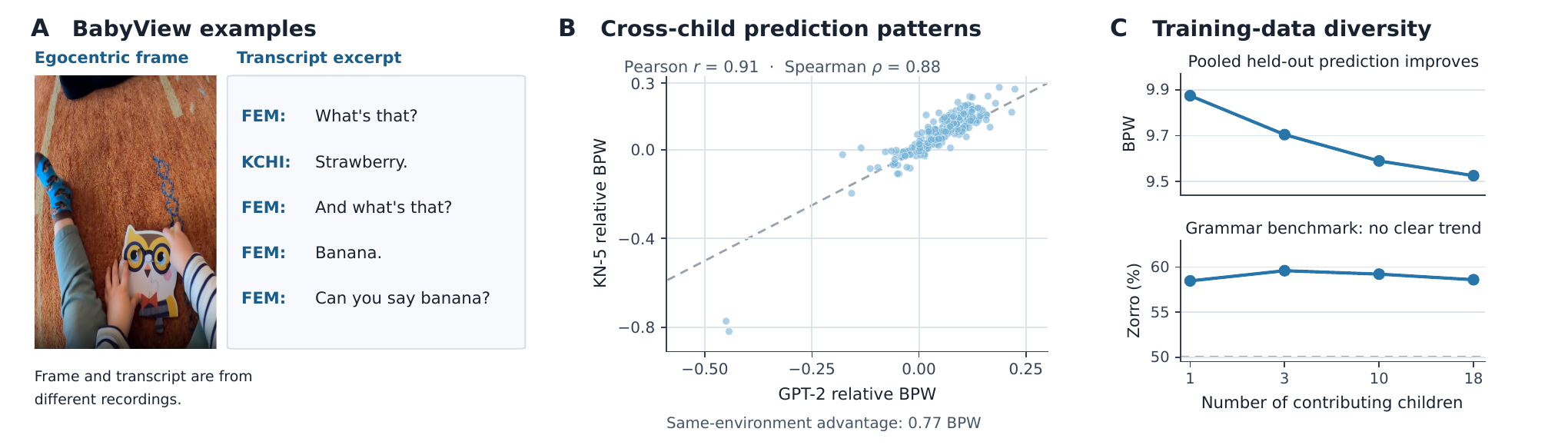}
\caption{\textbf{BabyView examples and two central findings.} \textbf{(A)} A representative BabyView egocentric frame and a separate excerpt of automatically transcribed child--adult conversation. The egocentric frame and transcript excerpt come from separate BabyView recordings and are presented as illustrative examples. \textbf{(B)} Models specialize to local environments, and a 5-gram reproduces much of the cross-child pattern ($r=.91$ after excluding same-child comparisons). \textbf{(C)} With 100k training words, more contributing children improve prediction of pooled held-out data; Zorro (grammaticality) shows no clear trend. The dashed diagonal in (B) indicates equality between the adjusted model scores; the horizontal dashed line in (C) indicates theoretical chance performance on Zorro.}
\label{fig:overview}
\end{figure*}

\section{Introduction}\label{introduction}

Modern language models learn from enormous amounts of text. Human children, by comparison, are exposed to far less language during early development: roughly \(10^7\)--\(10^8\) words, compared with around \(10^{12}\)--\(10^{13}\) words used to train many current large language models (LLM) \citep{frank2023datagap, warstadt2023babylm}. This large ``data gap'' has motivated developmental-scale language modeling, including the BabyLM challenge \citep{warstadt2023babylm, hu2024babylm2}. BabyLM keeps the training data fixed and compares different learning methods. We take the complementary approach: we mostly keep the learner fixed and compare different kinds of training data. These models are not meant to reproduce how children learn. We instead use them to ask what properties of small language datasets make learning and generalization easier or harder. 

This data gap is not only about the number of words a learner receives. Two learners can see the same number of words but experience very different kinds of language. Children's environments differ in vocabulary, topics, speakers, routines, and linguistic structures.
These differences may matter most when training data are limited. For example, a model may perform well on a test set simply because the same words and constructions appeared frequently in its training data. We use distributional support and match to refer to these ideas: whether the language being evaluated is represented in the training data, and how similar the training and evaluation distributions are.

Naturalistic developmental data make it possible to study these questions directly. The BabyView dataset contains egocentric recordings from children's everyday environments, along with natural language transcripts \citep{long2024babyview}. Since many children were recorded using the same collection procedure, we can train separate models on different children's environments while keeping the broader data-collection setup consistent. 
Because densely recorded language from individual children is limited, however, our training sets are still smaller than children's estimated total exposure.
Thus, we pair BabyView with TinyDialogues \citep{feng2024tinydialogues}, a synthetic child-directed dialogue corpus that can be generated at larger scales.

We also compare our neural models with a simple count-based baseline. When training data are very small, strong prediction may not require complex neural representations: vocabulary overlap and short-range word patterns may already explain much of the performance. We therefore train a Kneser-Ney 5-gram model on each training set. This baseline serves two purposes. First, it tells us when a simple count model can match or outperform GPT-2 at our data scale. Second, it helps us ask whether effects that look like sophisticated environmental learning can instead be explained by relatively simple distributional statistics.

With this setup, we investigate three main questions. 1) How strongly do models specialize to individual children's language environments, and what explains that specialization? 2) At a fixed training budget, what changes when the data comes from more children? 3) How do learning curves and benchmark performance differ between natural and synthetic input as the amount of data increases?

Figure~\ref{fig:overview} highlights the dataset and our two most distinctive findings. Models develop large and persistent advantages for the environments they were trained on, much of which can be explained by lexical and other distributional differences. At a fixed data budget, mixing language from more children improves prediction of pooled held-out language more clearly than grammatical performance. We also examine how natural and synthetic inputs support grammatical and lexical-semantic learning as data increase, using count models as a control. 

\section{Related work}\label{related-work}

\paragraph{Developmental-scale language modeling.}\label{developmental-scale-language-modeling}

Recent work asks how language models can learn from datasets far smaller than those used in standard pretraining. The BabyLM challenge limits training data and evaluates models on common grammatical and semantic tasks \citep{warstadt2023babylm, hu2024babylm2}. Other work trains small LMs on child-directed or developmentally motivated data, including CHILDES-derived corpora \citep{huebner2021babyberta, portelance2023aoa, yedetore2023poor, hu2025production} and synthetic dialogue \citep{feng2024tinydialogues, zeng2026bilingual}. Single-child studies use transcribed input or grounded video and audio \citep{wang2023structure,vong2024grounded}; \citet{qin2024single} trained on three single-child datasets of roughly 0.2M tokens each. An earlier BabyView study examined scaling, differences among children's input, and developmental vocabulary outcomes on a subset of the current data \citep{feng2026babyscale}. Our study builds on this setting but uses matched data budgets, more training draws and seeds,
systematic cross-child comparisons, and additional tests of specialization, persistence, and diversity. Following a controlled-rearing approach \citep{warstadt2022controlled}, we mostly hold the model and training pipeline fixed while varying the input, rather than comparing learning methods on a fixed corpus.

\paragraph{Scaling and data efficiency.}\label{scaling-and-data-efficiency}

Scaling laws describe how LM performance changes as compute, model size, and training data increase. Across large ranges, loss often follows smooth relationships that can be approximated by power laws \citep{kaplan2020scaling, hoffmann2022training, muennighoff2023scaling}. Related work also examines how to allocate data between pretraining and retrieval \citep{singh2026memorize}. Most of this evidence, however, comes from datasets many orders of magnitude larger than what we study. At small data scales, factors such as optimization and unseen vocabulary or constructions may matter more. Our goal is to characterize scaling within the child-scale data regime, not to propose or test new scaling laws.

\paragraph{Input distribution, domain match, and variation across children.}\label{input-distribution-domain-match-and-variation-across-children}

Models perform better when their training data resembles the data they are evaluated on, and the mixture of training data can affect downstream performance \citep{gururangan2020dontstop, longpre2024pretrainer, moore2010intelligent}. Developmental research shows that children's language environments differ both in how much language they hear and in what that language contains, including differences in vocabulary and syntax \citep{rowe2012longitudinal, rowe2020input, anderson2021linking, masek2021beyond}. These findings motivate separating the amount of input from its composition and its match to the evaluation data.

\section{Experimental framework}\label{experimental-framework}

\subsection{Training corpora}\label{training-corpora}

\textbf{BabyView.} We use naturalistic transcripts from the 2026.1 release of the BabyView dataset, which contains egocentric recordings (along with natural language transcripts) of children's everyday environments. After preprocessing and filtering the English data, the available training data contains approximately 4.67M words. Our main scaling series uses nested random subsets of 25k, 50k, 100k, 200k, 400k, 800k, 1.6M, and 3.2M words, with three corpus draws at each data size. We also evaluate the full 4.67M-word corpus with one draw and three model seeds; this is shown separately from the replicated scaling series. For each draw, we randomly order the corpus using a fixed seed, and larger data sizes extend the smaller ones. Our uncertainty intervals therefore describe variation within the observed BabyView data, rather than variation across a hypothetical population of households.

\textbf{TinyDialogues.} As a synthetic comparison, we use child-directed dialogue generated for a two-year-old child (the \emph{age-2} condition) \citep{feng2024tinydialogues}. Its training sizes range from 25k to 12.8M words. Supplementary experiments vary the target ages of the generated dialogue: M1 combines ages 2, 5, and 10 under the same generation setup, while M2 also changes the generator source. Because M2 changes more than age composition, we do not use it to isolate an age effect (Appendix~\ref{app:agecomp}).

\textbf{Individual environments and mixtures.} To study specialization, we construct matched 100k-word training corpora for 16 children, with three corpus draws and three model seeds per child (144 runs). Each model is evaluated on held-out language from its own child's environment and the other children's environments. 
Of 18 available children, 16 have enough training data for a 100k-word single-child corpus. To study diversity, we vary the number of contributing children (\(k\in\{1,3,10,18\}\)) at a fixed 100k-word total budget, using the 16 eligible children for \(k=1\) and drawing multi-child mixtures from all 18 children for other values of $k$. We also perform a 250k-word replication as a robustness check. We use 100k words for the primary analysis because this budget allows matched single-child comparisons across 16 eligible children, whereas the 250k-word follow-up only includes six individual children and is thus treated as a secondary replication.

The three experiment families use different model and tokenizer configurations where required; we keep these separate throughout the analyses.

\subsection{Models and training}\label{models-and-training}

The controlled scaling experiments use GPT-2-style causal language models \citep{radford2019gpt2} at two sizes: 13.9M-parameter micro and 91.6M-parameter small. Each scaling condition uses a separate 7.5k-token BPE tokenizer trained on held-out data from that corpus. Models are trained for 20 epochs over the data, with checkpoints selected using validation loss. The primary checkpoint is the one with the lowest loss on a held-out selection split that is separate from all reported evaluation sets; we also report results from the final checkpoint as a secondary analysis. The individual-environment, persistence, and diversity analyses instead use GPT-2 micro with one shared 16k-token BPE tokenizer across all children (17.2M parameters), and use the final checkpoint for the main analysis.

We match training conditions by the number of sampled word occurrences, not optimization steps or compute. Since every model makes the same number of passes over its dataset, models trained on larger datasets receive more optimization steps. We call these comparisons fixed-recipe scaling and do not interpret them as compute-optimal learning curves. To check whether the low-data comparison with the count model depends on our training setup, we test alternative learning rates, dropout, weight decay, and training lengths from 100k to 1.6M words. For each corpus, model size, and data size, we choose the configuration with the best average validation loss across the three corpus draws, without using the held-out evaluation data, and confirm it with additional model seeds. Stronger regularization and longer training are selected in most cases, suggesting that the fixed recipe is not optimal across this range (Appendix~\ref{app:recipe}).

\subsection{Kneser-Ney 5-gram baseline}\label{kneser-ney-5-gram-baseline}

For each training corpus, we fit a Kneser-Ney 5-gram (KN-5) model \citep{kneser1995,chen1999} as a simple baseline based on local word statistics. KN-5 lowercases the text and conditions only within utterances, while GPT-2 can use longer context as well as speaker and separator tokens. The two models thus make different assumptions about the data, so we use this comparison as a diagnostic rather than as a perfectly matched architecture ablation.

KN-5 and GPT-2 handle unseen words differently. KN-5 maps all unseen words to a single \texttt{\textless{}unk\textgreater{}} symbol, while BPE must assign probability to the full spelling of each word. This can favor KN-5 when many held-out words are unseen during training. We hence base our primary crossover analysis on a vocabulary-aligned comparison, scoring both models on the same in-vocab held-out words. Results using the original closed-vocabulary scoring are reported in Appendix~\ref{app:kn5}.

\subsection{Evaluation}\label{evaluation}

\textbf{Held-out prediction.} We report content bits per word (BPW), where lower values mean better prediction. Our main cross-corpus scaling curves use held-out text from the corresponding corpus. Because the evaluation text therefore differs across corpora, we do not interpret absolute BPW differences between corpora as controlled corpus effects. When direct comparisons are needed, we instead evaluate models on the same text.

\textbf{Grammatical benchmarks.} Zorro \citep{huebner2021babyberta} contains 46k grammatical minimal pairs covering 13 phenomena. Our primary evaluation compares the summed log-probabilities of the two sentences in each pair. Main figures plot Zorro accuracy. We also report a per-token-normalized Zorro sensitivity analysis and BLiMP \citep{warstadt2020blimp} as a supporting benchmark. Because effects differ heavily across phenomena, we refer to these results as ``grammatical benchmark performance'' rather than treating aggregate accuracy as a direct measure of grammatical competence.


\textbf{Lexical semantics.} We evaluate how well models capture relationships between word meanings using five word-similarity benchmarks \citep{rubenstein1965rg,finkelstein2002ws353,hill2015simlex,gerz2016simverb,bruni2014men}, in which humans rated how similar or related pairs of words are. For each pair, we compute the cosine similarity between the model's word representations and measure how well they agree with human ratings using Spearman correlation. We extract word representations from the final transformer layer, averaging across subtokens when necessary. This extraction method is fixed rather than choosing the best-performing layer on each test set; absolute scores may differ under other methods (Appendix~\ref{app:wordsim-sensitivity}). Cross-corpus comparisons use identical benchmark pairs. Developmental outcomes are supplementary (Appendix~\ref{app:developmental}). We use WordSim in the corpus-scaling comparisons; the fixed-budget diversity experiment instead focuses
on held-out prediction and grammaticality.

\subsection{Statistical analysis}\label{statistical-analysis}

Training runs are replicated across corpus draws and model seeds. Scaling intervals use a two-stage bootstrap over corpus draws and seeds; individual-environment and persistence analyses use child-cluster bootstraps and permutation tests; diversity analyses use HC1 intervals and label permutations. For the low-data scaling analysis, we test whether a standard power law adequately describes the observed curves using criteria specified before examining the results. Full 
details are in Appendix~\ref{app:stats}.

\section{Individual environments leave strong distributional signatures}\label{individual-environments-leave-strong-distributional-signatures}

\begin{figure*}[t]
\centering
\includegraphics[width=.96\textwidth]{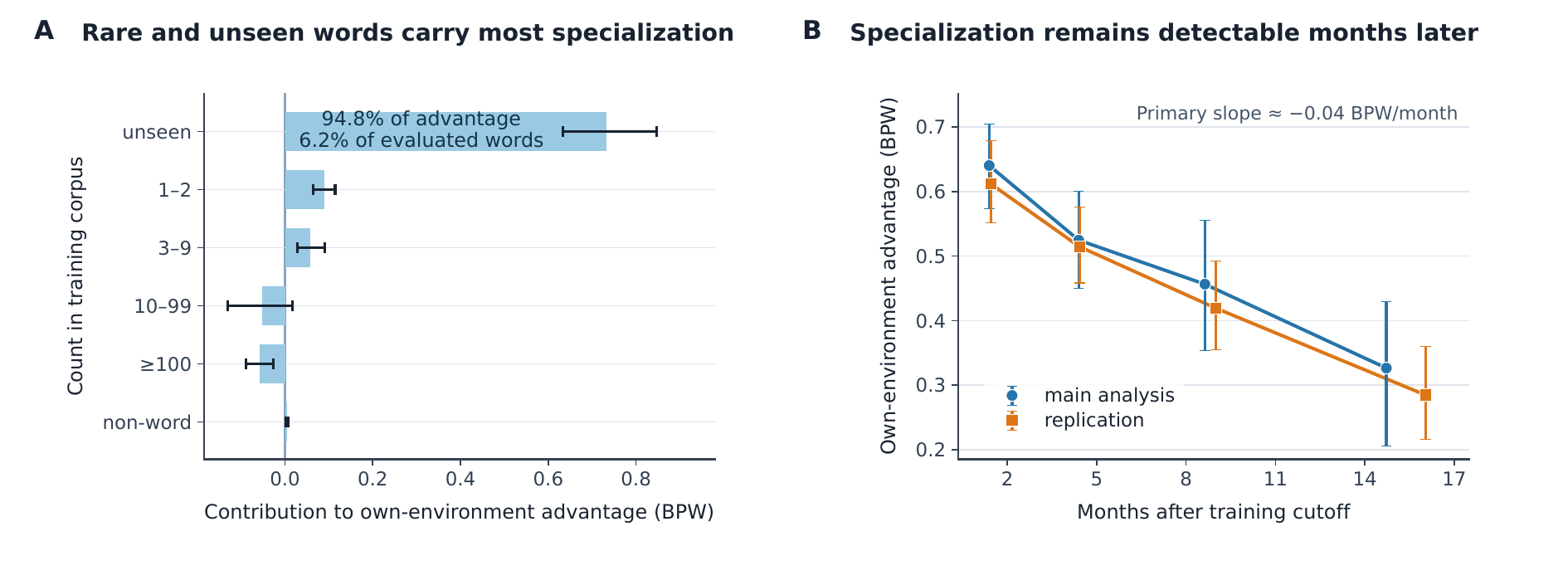}
\caption{\textbf{Specialization is concentrated in rare vocabulary and remains detectable months later.} \textbf{(A)} Words unseen in the source child's 100k-word training corpus account for 94.8\% of the same-environment advantage while comprising 6.2\% of evaluated words. \textbf{(B)} The advantage remains on later recordings but becomes weaker with time.}
\label{fig:specialization}
\end{figure*}

\subsection{Own-environment prediction advantage}\label{models-preferentially-predict-their-own-environments}

We first ask whether models trained on equal amounts of language become specialized to particular children's environments. To test this, we train 144 models (16 children $\times$ 3 corpus draws $\times$ 3 seeds) on matched 100k-word samples and evaluate each on held-out language from every child's environment. Across these comparisons, models show a strong same-environment advantage (Figure~\ref{fig:overview}B). On average, they achieve 0.77 BPW lower loss on held-out language from the environment they were trained on than language from other children's environments (95\% interval: [0.62, 0.92]). All 16 children show this advantage. By contrast, differences across children on Zorro are small relative to variation from corpus sampling and model runs (variance ratio 0.13). Children's environments thus differ strongly in how well they match each model's training data, without corresponding evidence that some environments provide generally better grammatical input.

\subsection{Beyond the child's own speech}\label{specialization-extends-beyond-the-target-childs-speech}

One possibility is that models specialize mainly by learning
the child's own way of speaking. To test this, we first
evaluate models using only speech from other people in the
child's environment. The same-environment advantage remains
essentially unchanged (0.775 BPW versus 0.769 BPW when
evaluating all speech). This shows that specialization also
extends to the language spoken by other people around the
child.

However, these models were still trained on the child's
full dialogue. In a separate experiment, we therefore remove
the child's own speech from training and compare this with
removing the same amount of speech at random. Removing the
child's speech specifically leads to worse held-out
prediction (+0.26 BPW across 10 children), while Zorro
performance is essentially unchanged
(Appendix~\ref{app:childspeech}). Together, these results
show that specialization extends beyond predicting the
child's own speech, even though that speech also provides
useful information during training.


\subsection{Lexical and distributional differences}\label{most-of-the-advantage-is-lexical-and-distributional}

The same-environment advantage is highly concentrated in particular words (Figure~\ref{fig:specialization}A). About 95\% of the effect comes from words that were unseen in the model's own training data, even though these words make up only about 6\% of the evaluation set. Cross-child prediction also becomes worse as vocabulary and lexical-distribution mismatch increase. Unseen words can include proper names, household-specific terms, rare objects, and transcription artifacts, and we do not fully separate these sources. We therefore interpret the specialization as mainly lexical and distributional, rather than as evidence that each child's environment produces a different global grammar.

\subsection{5-gram reproduces cross-child pattern}\label{a-5-gram-reproduces-much-of-the-cross-child-structure}

A simple KN-5 model reproduces much of the same cross-child structure. Even after excluding the 16 same-child comparisons, the KN-5 and neural matrices correlate strongly (\textbf{$r=.91$}) across the remaining 240 comparisons (Figure~\ref{fig:overview}B; full matrices in Appendix~\ref{app:crosschild}). KN-5 and GPT-2 hence largely agree about which children's environments are more or less compatible. This suggests that much of the cross-child pattern can be explained by local distributional statistics without requiring a neural explanation. We do not compare the raw KN-5 and GPT-2 effect sizes directly because KN-5 uses a closed vocabulary and the specialization effect is concentrated in unseen words; our main result is the shared pattern across environments.

\begin{figure*}[t]
\centering
\includegraphics[width=.96\textwidth]{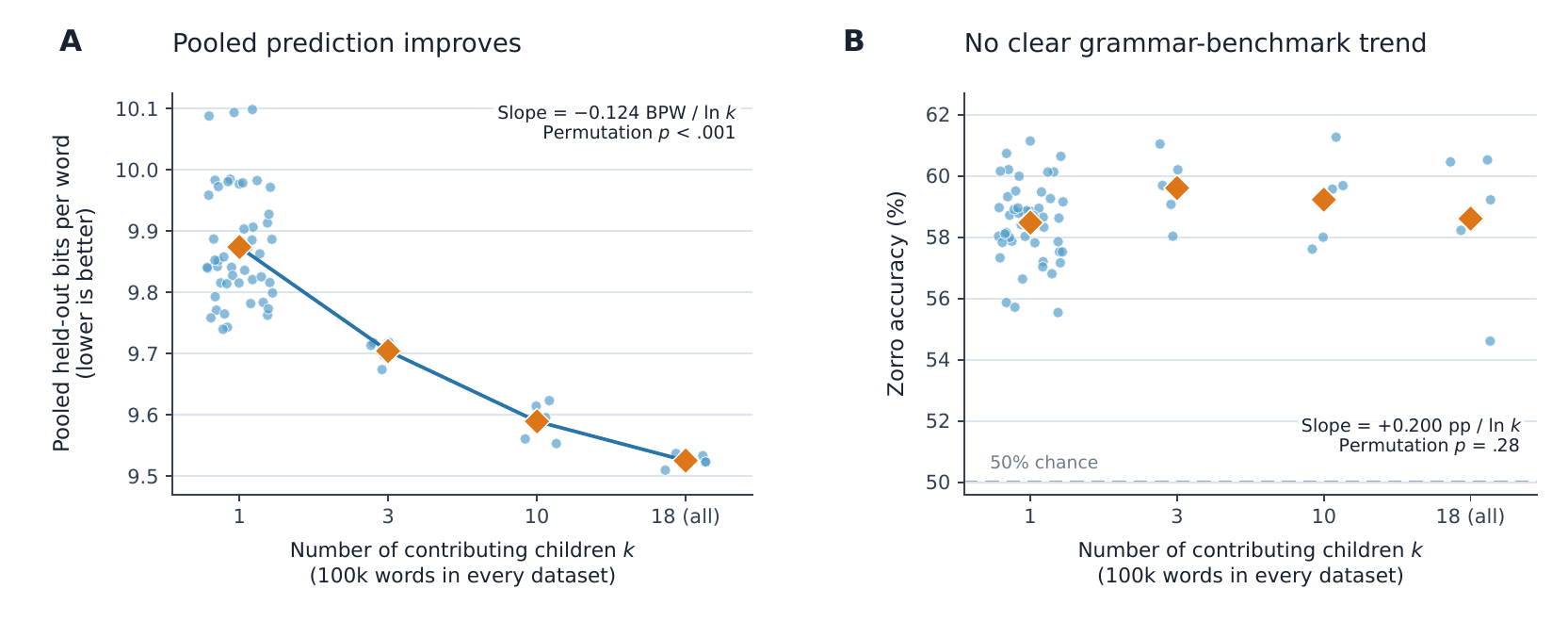}
\caption{\textbf{At a fixed word budget, language from more children improves pooled held-out prediction more clearly than grammatical-benchmark performance.} Each point is one 100k-word training dataset (mean across three model seeds); diamonds show the mean for each number of contributing children. \textbf{(A)} Pooled held-out loss decreases as the training set has more children (slope $-0.124$ BPW per $\ln k$; permutation $p<.001$). \textbf{(B)} Zorro shows no clear monotonic trend at the same budget ($p=.28$). 
The dashed line is theoretical 50\% chance on Zorro.}
\label{fig:diversity}
\end{figure*}

\subsection{Specialization persists for months}\label{specialization-persists-but-attenuates-over-time}

The specialization effect is not limited to language recorded around the same time as the training data. We train models on each child's input up to a cutoff date and evaluate them on speech from other people in that child's environment at later times. Even 6--12 months later, the mean same-environment advantage remains $\sim$0.46 BPW. The effect becomes weaker over time, decreasing by $\sim$0.04 BPW/month; a separate chronological-prefix analysis shows a similar decline (Figure~\ref{fig:specialization}B). Because time and child age increase together, this decline should not be interpreted as a causal effect of time itself. The result shows that environment-specific differences remain detectable months later but become weaker as the child's language environment changes. More details in Appendix~\ref{app:persistence}.

\subsection{Distributional differences \& specialization}\label{controlled-heterogeneity-is-sufficient-to-induce-specialization}

Finally, we test if differences between environments are sufficient to produce this kind of specialization in a controlled synthetic setting. As we make the synthetic source distributions more different from one another, the same-environment advantage increases steadily. At similar levels of distributional difference, synthetic data can produce specialization of a similar overall size to BabyView, although BabyView's effect is much more concentrated in unseen vocabulary. Distributional differences alone can thus produce substantial specialization, but the synthetic manipulation does not reproduce all of the lexical structure found in the naturalistic data. The amount of distributional difference also does not fully predict the advantage across different synthetic manipulations. Experimental details and comparisons are in Appendix~\ref{app:synthetic-specialization}.

\section{Diversity improves pooled prediction more clearly than grammar}\label{diversity-improves-pooled-distribution-prediction-more-clearly-than-grammatical-benchmark-performance}

The specialization results show that training on a particular environment helps a model predict that same environment. We next ask the complementary question: if the total amount of training data stays fixed, what changes when that data comes from more children's environments? We construct 100k-word training sets containing language from 1, 3, 10, or 18 (all available children) environments. As the number of contributing children increases, prediction of the pooled held-out data improves substantially (Figure~\ref{fig:diversity}A). BPW decreases by about 0.124 for each unit increase in $\ln k$.

This improvement closely follows how well the training data matches the pooled evaluation data. In a leave-one-dataset-out analysis, unigram Jensen-Shannon divergence \citep{lin1991jsd} explains more variation in performance than the number of contributing children alone ($R^2=.93$ versus $.67$; Appendix~\ref{app:diversity}). In other words, mixtures that more closely resemble the pooled target distribution are easier for the model to predict. We do not interpret this relationship as causal, nor test transfer to a completely new child.

The grammar benchmarks show a different pattern. At 100k words, Zorro shows no clear trend with the number of contributing children ($p\approx .28$), and per-token Zorro and BLiMP are likewise null. Thus, mixing more environments clearly improves prediction of pooled held-out language but does not produce a similarly clear grammatical-benchmark improvement at 100k words (Figure~\ref{fig:diversity}B).

The 250k-word robustness check is less clear-cut: per-token Zorro is positive ($p=.044$), while the primary Zorro result is weaker ($p=.085$) and BLiMP remains null. We therefore do not claim that diversity never helps grammatical generalization (Appendix~\ref{app:diversity}). The narrower conclusion is that its benefit for pooled prediction is clear at the primary 100k-word budget, while its effect on grammatical benchmarks is smaller and less consistent.

\section{Learning across data sizes and sources}\label{learning-trajectories-at-child-scale}

\begin{figure*}[t]
\centering
\includegraphics[width=.96\textwidth]{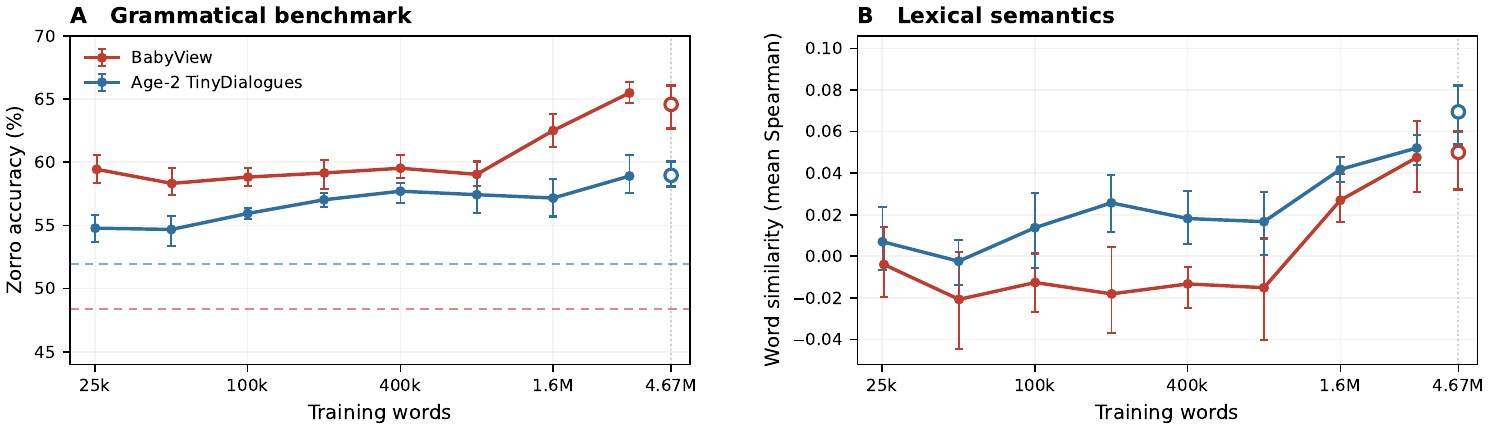}
\caption{\textbf{Scaling of Zorro and WordSim performance (GPT-2 small).}
\textbf{(A)} Zorro accuracy (50\% is chance; color-matched
dashed lines show random-initialization baselines: 48.4\% for
BabyView and 52.0\% for age-2 TinyDialogues).
\textbf{(B)} Mean WordSim Spearman correlation across five datasets
(the same 7,917 word pairs for both corpora).
Lines connect the shared 25k--3.2M-word conditions
(three corpus draws $\times$ three seeds each), with pointwise
95\% two-stage bootstrap intervals.
Unconnected hollow markers show 4.67M-word results:
full BabyView (one draw, three seeds; seed-only range) and
size-matched age-2 TinyDialogues (three draws; bootstrap interval).
Both use 20 training epochs and validation-selected checkpoints.
Appendix~\ref{app:micro-scaling} shows GPT-2 micro and larger
age-2-only conditions; Appendix~\ref{app:agecomp}
mixed-age comparisons; Appendix~\ref{app:wordsim-sensitivity}
the common-word-pair analysis.}
\label{fig:rawscaling}
\end{figure*}

\subsection{Scaling differs by data source}\label{naturalistic-and-synthetic-input-produce-different-grammatical-profiles}

As training data increase, BabyView and age-2 TinyDialogues show different patterns of performance on grammatical and lexical-semantic benchmarks (Figure~\ref{fig:rawscaling}). Across their shared data sizes, BabyView-trained models generally score higher on Zorro, especially at the smallest sizes and again from 1.6M to 3.2M words. The advantage is weaker around 200k--800k words under per-token scoring and on BLiMP. The lexical-semantic comparison points in a different direction: TinyDialogues often performs better on word-similarity benchmarks. Figure~\ref{fig:rawscaling} evaluates the same benchmark word pairs for both corpora (GPT-2 micro results: Appendix~\ref{app:micro-scaling}). As a more restrictive check, the fixed common-word-pair analysis gives a BabyView-minus-age-2 WordSim difference of $-0.057$ for GPT-2 small across 800k--3.2M words (Appendix~\ref{app:wordsim-sensitivity}). At 25k words, BabyView-trained GPT-2 small scores 59.4\% on Zorro versus 48.4\% at matched random initialization; the improvement is concentrated in lexical and local-word phenomena, not broad agreement (Appendix~\ref{app:lowdata-zorro}). This is a task-specific comparison, not a claim that either corpus supports globally better language learning.

\subsection{Grammar differences by phenomenon}\label{grammatical-differences-by-phenomenon}

Overall Zorro accuracy hides large differences among grammatical phenomena. For GPT-2 small across 800k--3.2M words, negative-polarity-item (NPI) licensing accounts for much of the BabyView advantage (about 32 percentage points on that phenomenon; Figure~\ref{fig:zorro-phenomena} in
Appendix~\ref{app:kn5}). BabyView also performs better on binding, island effects, and irregular forms, while TinyDialogues does better on several other categories, most consistently quantifiers. These results do not show that BabyView is universally better grammar training data.

The phenomenon-level comparison with KN-5 sharpens this distinction (Appendix~\ref{app:kn5}). The count model shows a smaller overall BabyView advantage on Zorro and does not reproduce the neural pattern; most notably, the NPI difference reverses in favor of age-2 TinyDialogues. The observed neural NPI advantage is concentrated in the \textit{only/even} subset, and differences in licensor frequency do not explain it. We do not have a full mechanistic explanation for this, nor establish a general grammatical advantage for either input source from these results.

\subsection{Count models stay competitive at low data}\label{count-models-remain-competitive-at-the-smallest-budgets}

As a secondary diagnostic, we compare GPT-2 with KN-5 on held-out prediction. With a fixed 20-epoch training setup and vocabulary-aligned evaluation, GPT-2 small first overtakes KN-5 between 800k and 1.6M words on BabyView and between 200k and 400k on age-2 TinyDialogues. Tuning the neural training setup using validation data moves the BabyView crossover earlier, to 400k--800k words, but does not change the age-2 small-model range. This shows that simple local statistics remain competitive when data are scarce, while the exact crossover depends on training choices (Appendix~\ref{app:recipe}; Figure~\ref{fig:count-crossover}).

The fixed-training learning curves improve overall with data, but a single standard power law does not pass our pre-specified fit checks on this low-data range. This is a limited statement about the tested models and fitting procedure, not a rejection of power-law scaling more generally (Appendix~\ref{app:powerlaw}).

\section{Discussion}\label{discussion}

Our results suggest that the gap between LM training data and children's experience is not just about quantity. At small data sizes, it matters which words and constructions are present and how well they match the language being evaluated. Models specialize strongly to individual children's environments, while including more children's environments improves prediction of pooled held-out language without a similarly clear benefit for grammatical benchmarks. These findings highlight how different properties of language input can affect prediction and generalization in different ways.

Although some degree of specialization is expected from differences between training and evaluation distributions, its magnitude and structure in individual children's language environments are empirical questions. We find that these environments leave large and persistent signatures in model predictions, yet much of this specialization is concentrated in a small fraction of evaluated words and reproduced by a simple count model. At the same time, differences among children in grammatical-benchmark performance are comparatively small, suggesting that environmental match and broader linguistic generalization need not vary together.

Our scaling results further show that different sources
of training data can support different aspects of language
learning. BabyView-trained models generally performed better
on Zorro, while models trained on synthetic TinyDialogues
often performed better on word-similarity benchmarks.
The grammatical differences varied substantially
across phenomena, 
and the 5-gram baseline did not reproduce the neural model's pattern, suggesting that
simple local word statistics cannot explain all the
differences between training sources. These
findings are specific to the datasets, models, and
benchmarks studied here, and do not establish that either
training source is generally better for learning.

Children face the challenge of learning general language patterns from small, highly variable, and often household-specific samples. Yet they still generalize beyond their immediate environments. One possibility is that such learning requires stronger inductive biases or regularization than the models studied here use, together with social, sensory, and multimodal information. Our experiments do not directly test this explanation, but they help identify the distributional differences a theory of child language learning must address.

These findings should not be read as showing that GPT-2 learns language in the same way children do. The value of child-scale data is that it allows us to study realistic variation in language input at a scale where vocabulary overlap and specialization are easy to observe. Future work could test transfer to entirely unseen children, separate names and transcription artifacts from other unfamiliar words, compare larger or compute-matched models, and extend BabyView modeling from transcripts to grounded video and audio.

\section{Conclusion}\label{conclusion}

At small training-data sizes, language-model performance depends not only on how many words a model sees, but on which environments those words come from. Models specialize strongly to individual children's language environments, especially in vocabulary and other distributional features, and that specialization remains detectable months later. Including more children's environments at a fixed training budget improves prediction of pooled child-language data without a similarly clear gain on grammatical benchmarks. Naturalistic BabyView and synthetic dialogue also support different grammatical and lexical-semantic learning patterns as training data increase; simple count models help identify which patterns can be explained by local statistics.

Together, these findings show why data quantity alone is not enough to describe language learning with small datasets. The content and diversity of the input, and how well it matches the language the model is asked to predict, also matter.


\newpage
\section*{Limitations}\label{limitations}

Our conclusions are limited to the models, datasets, and evaluations studied here. BabyView samples only part of children's language and multimodal experience, and dense per-child recordings are scarce: even our pooled 4.67M-word corpus is much smaller than estimates of children's cumulative exposure during early development, which can reach tens of millions of words. Our 100k-word individual-child experiments are smaller still. These results should therefore not be interpreted as modeling the full input available to a child. Further, GPT-2-style models are not cognitive models of children, and benchmarks such as Zorro, BLiMP, and WordSim measure particular model behaviors rather than human linguistic knowledge.

More broadly, we use these language models to study the effects of training data on general statistical learners so that these findings can be compared to the performance of children learning language. We do not believe that GPT-2 is a good model of how children themselves learn language, only that a comparison between the two might be informative.

Our scaling comparisons are specific to the training recipes, models, and data sizes tested; the largest synthetic training sets have no matched BabyView counterparts, and TinyDialogues reflects its generation setup. The diversity experiment measures prediction of pooled held-out data from the observed children, not transfer to a new household.

\section*{Ethical Considerations}\label{ethical-considerations}

BabyView contains naturalistic recordings of children and was collected under approved consent and ethics procedures, as described in the original dataset paper \citep{long2024babyview}. We use the existing dataset for secondary analysis, collect no new participant data, and do not reproduce identifying participant information. The example in Figure~\ref{fig:overview}A is an automatically generated transcript from a separate held-out recording. Further, TinyDialogues is an entirely synthetic dataset. Although our work here is purely basic research, we hope that a better understanding of children's language learning -- derived in part from studies comparing children and language models -- might eventually lead to better language and speech interventions for children.

\section*{AI Use Statement}
We used generative AI to assist with experimental planning and design, implementation, analysis, literature search, and for writing assistance, including rephrasing and improving clarity of author-written text. We reviewed and verified AI-assisted code, results, citations, claims, and text against primary sources and experimental artifacts. The authors directed the research and take responsibility for the final content, including text, claims, or artifacts produced with the assistance of generative AI.

\section*{Acknowledgements}

We gratefully acknowledge RunPod, Modal, and NVIDIA’s Academic Grant Program for providing the compute resources that enabled this work.

\bibliography{references}

\newpage
\appendix
\section{Training-Recipe Sensitivity}\label{app:recipe}
The fixed-recipe analyses use learning rate $3\times10^{-4}$, dropout 0.1, weight decay 0.1, and 20 epochs. We also evaluated seven alternative configurations varying learning rate ($10^{-4}$, $3\times10^{-4}$, $10^{-3}$), dropout (0.1 or 0.3), weight decay (0.1 or 0.5), and training length (20 or 40 epochs), at 100k--1.6M words for both BabyView and age-2 TinyDialogues and both model sizes. The screening used three corpus draws with seed 0 (480 runs); selected configurations were confirmed with two additional model seeds (120 runs). Selection minimized mean loss across draws on a held-out selection split, without using the reported evaluation sets. The strongest-regularization, longer-horizon configuration (learning rate $3\times10^{-4}$, dropout 0.3, weight decay 0.5, 40 epochs) was selected in 14 of 20 cases; the original fixed configuration was selected in none. Consequently, this bounded grid establishes recipe sensitivity but not a global training optimum.

\begin{table}[ht]
\centering
\small
\setlength{\tabcolsep}{3pt}
\begin{tabular}{lll}
\toprule
Corpus/model & Fixed setup & Tuned setup \\
\midrule
BabyView micro & 800k--1.6M & 400k--800k \\
BabyView small & 800k--1.6M & 400k--800k \\
Age-2 TD micro & 400k--800k & 200k--400k \\
Age-2 TD small & 200k--400k & 200k--400k \\
\bottomrule
\end{tabular}
\caption{Vocabulary-aligned KN-5/GPT-2 crossover ranges at the primary checkpoint. Tuning moves the crossover one tested data size earlier in three of the four comparisons.}
\label{tab:recipe}
\end{table}

\begin{figure*}[t]
\centering
\includegraphics[width=.94\textwidth]{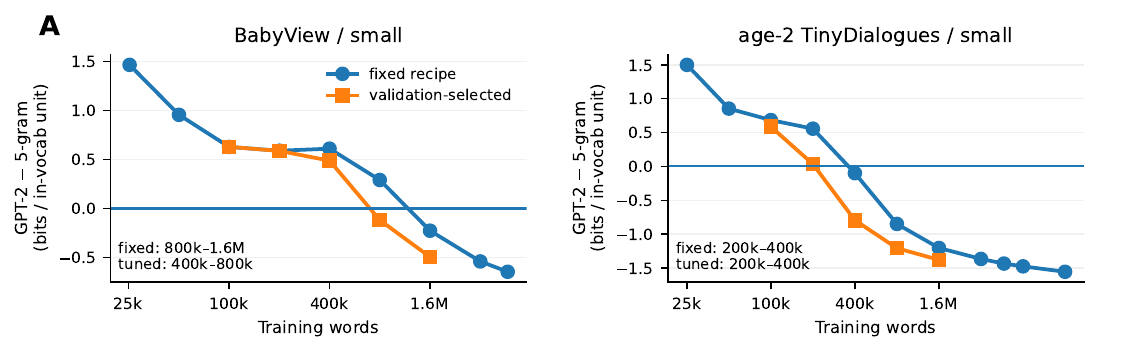}
\caption{GPT-2 small versus KN-5 on the same held-out in-vocabulary units. Positive values mean KN-5 predicts the evaluated units better; negative values mean GPT-2 is better.}
\label{fig:count-crossover}
\end{figure*}

The crossover ranges are shown in Table~\ref{tab:recipe} and illustrated for GPT-2 small in Figure~\ref{fig:count-crossover}. A replay of the fixed setup fell within the prespecified 0.005-nat tolerance in 54 of 60 cases (median absolute difference 0.00086 nats; maximum 0.0141), meeting the continuation rule. 

\section{Power-Law Fit Diagnostics}\label{app:powerlaw}
We fitted $L(N)=L_{\infty}+AN^{-\beta}$ to held-out loss at eight sizes (25k--3.2M words), weighting by variance estimates from three corpus draws. Before inspecting results, we specified five criteria for an adequate fit: (A1) parameter range, precision and agreement across starting values; (A2) fit relative to log-linear and log-quadratic alternatives, assessed with the small-sample corrected Akaike information criterion (AICc; \citealp{hurvich1989aicc}); (A3) weighted fit and residual patterns; (A4) leave-one-data-size-out predictive accuracy; and (A5) consistent, finite fits across the three draws.

The numerical thresholds were $\beta\in[0.05,1.5]$ with interval inside $[0.01,2.0]$, interval half-width at most $0.5\hat\beta$, and starting-value estimates agreeing within 0.001 (A1); $\Delta\mathrm{AICc}(\mathrm{power}-\mathrm{loglinear})\leq-2$ and $\Delta\mathrm{AICc}(\mathrm{power}-\mathrm{logquadratic})\leq 2$ (A2); weighted $R^2\geq0.98$, largest absolute standardized residual at most 2.5, and same-sign residual runs at most 4 (A3); at least 7 of 8 left-out sizes predicted within two standard deviations and mean absolute error at most $1.1$ times that of the log-quadratic comparator (A4); and three non-boundary per-draw estimates consistent with the pooled interval (A5).

No core BabyView or age-2 TinyDialogues fit passed the full criteria for either model size at the primary checkpoint. For example, BabyView micro had weighted $R^2=0.965$, and AICc was 71.9 for the power law versus 58.5 for the log-linear alternative; representative leave-one-size-out analyses included only 0--2 of eight sizes within two standard deviations. Thus the standard power law did not adequately summarize these observed curves under this stringent test. This does not exclude power-law relationships under other training setups or fitting procedures. 

\section{KN-5 Vocabulary Alignment and Grammar Controls}\label{app:kn5}
KN-5 is a lowercased, utterance-level word model with a closed vocabulary; unseen words map to an unknown-word symbol. GPT-2 instead models their full byte-level spelling and may condition on a longer context. The primary crossover comparison therefore sums both models' prediction costs over identical held-out units present in the corresponding KN-5 training vocabulary (bits per in-vocabulary unit). This reduces the KN-5 margin, but all four primary-checkpoint crossover ranges are unchanged. At the final checkpoint, BabyView small is the exception: its original closed-vocabulary crossover lies between 1.6M and 3.2M words, while its vocabulary-aligned crossover lies between 800k and 1.6M.

\begin{figure*}[t]
\centering
\includegraphics[width=.98\textwidth]{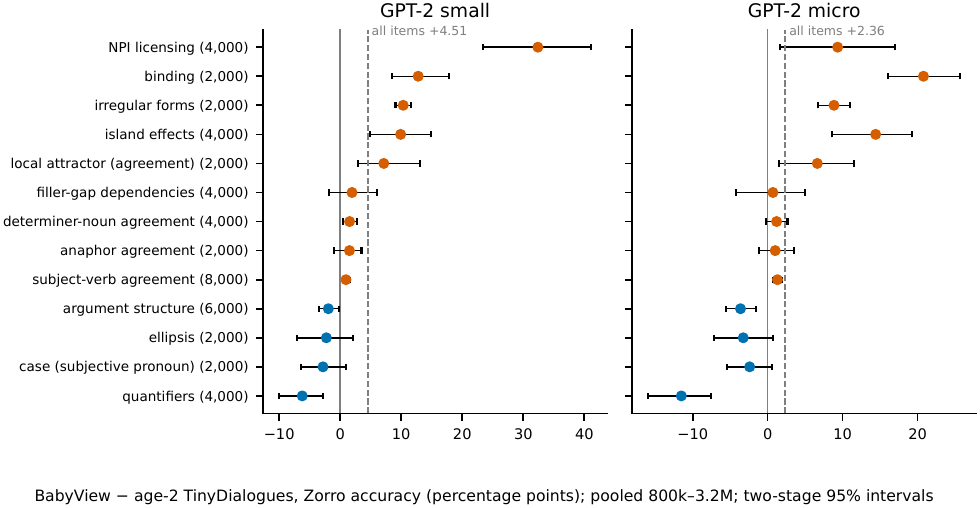}
\caption{Zorro accuracy differences by grammatical phenomenon (BabyView minus age-2 TinyDialogues) for GPT-2 small and micro, pooled across 800k--3.2M training words. Positive values favor BabyView. Error bars are pointwise 95\% two-stage bootstrap intervals across three corpus draws and three model seeds per arm; the overall difference is $+4.51$ percentage points for small and $+2.36$ for micro.}
\label{fig:zorro-phenomena}
\end{figure*}

On Zorro, the pooled BabyView-minus-age-2 TinyDialogues contrast was about $+1.69$ percentage points for KN-5 (95\% interval $[+0.34,+2.96]$), versus about $+4.51$ for the neural models. For negative-polarity-item (NPI) licensing the comparison instead reverses: about $-13.9$ points for KN-5, versus $+32.4$ for GPT-2. An item-level lexical-frequency explanation was not identifiable in the planned test because the intended predictor barely varies across the relevant NPI items; the measurable neural effect is concentrated in the \emph{only/even} subset. Thus the 5-gram comparison is an informative control, not an architecture-matched ablation or a full explanation of grammatical-benchmark differences. Likewise, KN-5 versus neural own-environment advantage magnitudes should not be compared directly under closed-vocabulary scoring. At the seven data sizes where the main Zorro interval excludes zero, per-token scoring agrees in direction at 7/7 sizes for each model; BLiMP agrees at 6/7 for small and 7/7 for micro, with weaker BabyView advantages around 200k--800k words. The full 13-phenomenon breakdown is shown in Figure~\ref{fig:zorro-phenomena}.

\section{Cross-Child Specialization Checks}\label{app:crosschild}
\begin{figure*}[t]
\centering
\includegraphics[width=.94\textwidth]{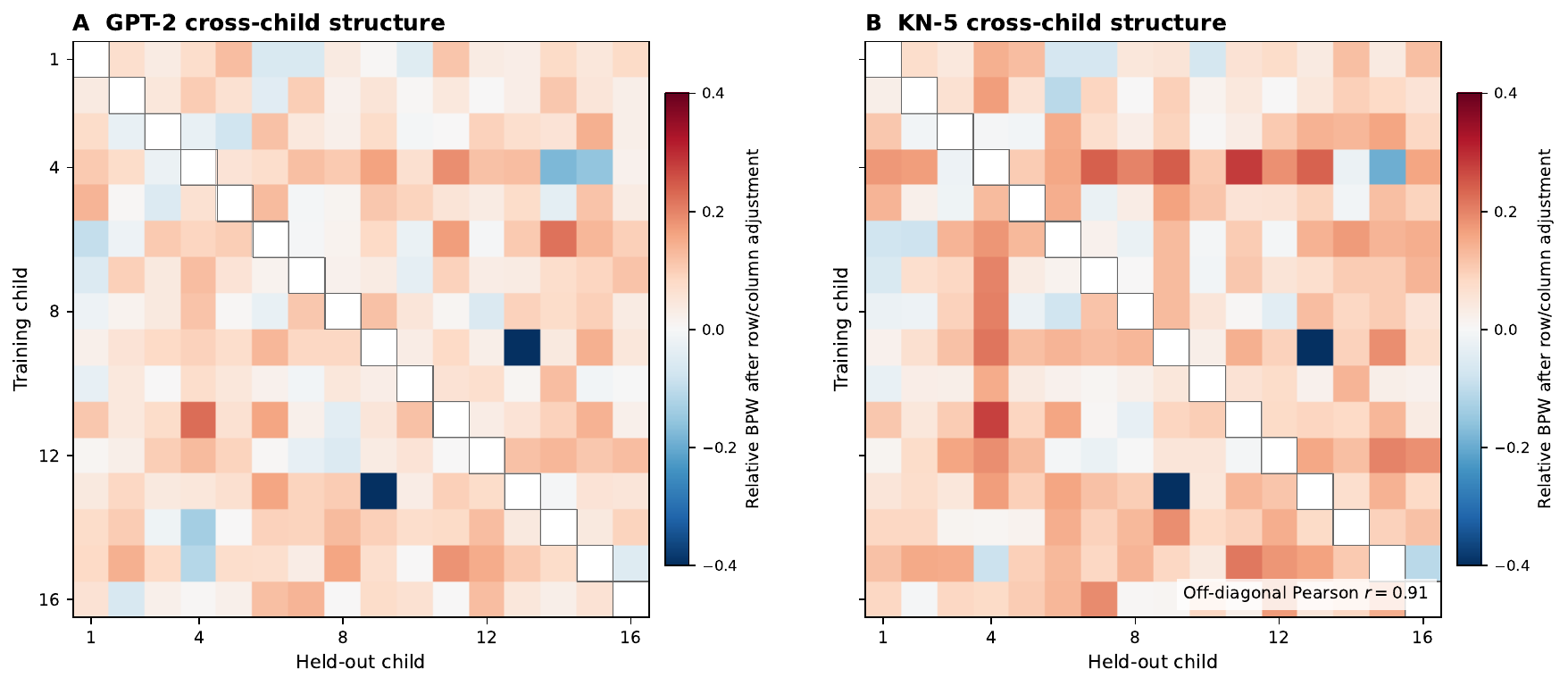}
\caption{Cross-child held-out prediction patterns for GPT-2 and KN-5 after the same row/column adjustment. Diagonal entries are masked and colors are clipped to $[-0.40,0.40]$ BPW to reveal the remaining 240 comparisons, whose patterns correlate at $r=.91$. The color clipping affects display only, not the correlation.}
\label{fig:crosschild-heatmaps}
\end{figure*}
Figure~\ref{fig:crosschild-heatmaps} shows the matched neural and count-model matrices. The overall own-environment advantage was 0.769 BPW (95\% interval $[0.623,0.921]$; all 16 children positive) for the all-speech comparison. In the non-target-speech evaluation, the own-environment advantage was approximately 0.775 BPW. The lexical decomposition and off-diagonal KN-5 correlation reported in the main text use this non-target-speech evaluation, not the all-speech matrix. Unseen words accounted for approximately 94.8\% of the advantage, despite constituting 6.2\% of evaluation words; these summaries describe the observed evaluation split and do not separate proper names, household-specific terms, and possible transcription artifacts. The neural and 5-gram matrices correlated at $r=.908$ after excluding the 16 same-child comparisons ($n=240$), and $r=.968$ across all 256 comparisons, following the same double-centering. The latter magnitude supports agreement in pattern, not agreement in raw loss scale. Further, the off-diagonal relationship in Figure \ref{fig:overview}B is not driven solely by extreme observations: Spearman correlation is $\rho=.883$, and removing the two most extreme reciprocal comparisons leaves Pearson $r=.864$. Recomputing the row- and column-centered matrices after excluding each child in turn yields correlations ranging from $.862$ to $.923$.

\section{Statistical Procedures and Analysis-Plan Provenance}\label{app:stats}
The replication unit for scaling is one training run (a corpus draw and model seed). Every partial-size corpus draw is a nested trajectory across data sizes: the two-stage bootstrap resamples whole trajectories, then model seeds within the selected draws (2,000 resamples for the main curves). The reported intervals are pointwise 95\% intervals conditional on the observed BabyView and TinyDialogues training corpora. 
Because we have only three corpus draws, our bootstrap uncertainty estimates are limited in precision. We also do not correct for multiple comparisons, so the number of confidence intervals excluding zero should be interpreted cautiously rather than as strong evidence for multiple separate effects. The full-corpus 4.67M BabyView anchor has one fixed corpus draw and three seeds, hence its displayed range reflects seeds only; the size-matched age-2 TinyDialogues anchor has three draws and a two-stage interval. Neither anchor should be confused with the repeatedly drawn 25k--3.2M series.

For individual-environment and persistence analyses, we use child-cluster bootstrap intervals and target-label permutation tests, as specified for each analysis; the persistence interval uses 10,000 child-cluster bootstrap resamples. Diversity uses HC1 intervals and Monte-Carlo label permutations over the 63 fixed-budget mixture datasets. The original project analysis plan designated within-child quantity, diversity, between-child heterogeneity, age of acquisition, and child-specific vocabulary as primary. The scaling, age-composition, lexical-locus, persistence, synthetic-heterogeneity, and diversity-explanation analyses were planned later, but their rules were recorded before the corresponding results were examined. These later analyses are not retroactively presented as original primary tests. 

\section{Persistence: Full Analysis and Replication}\label{app:persistence}
\begin{figure*}[t]
\centering
\includegraphics[width=.93\textwidth]{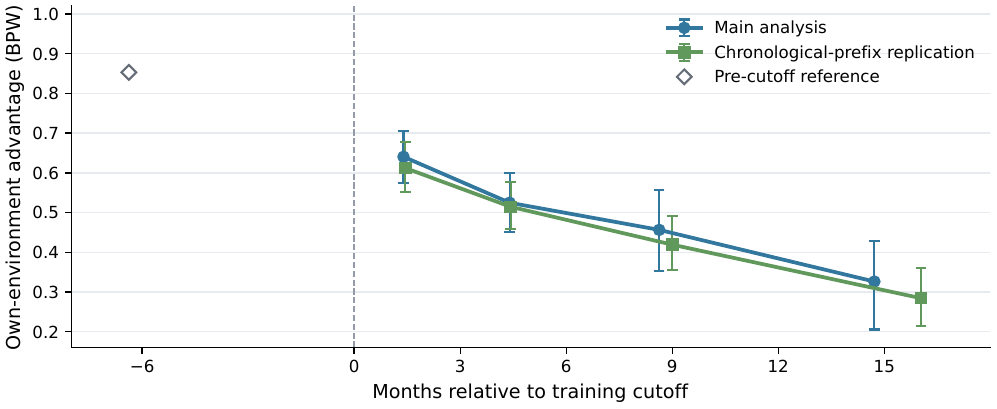}
\caption{Own-environment prediction advantage over time in the main analysis and an independent chronological-prefix replication. Points show averages for future evaluation windows and error bars show 95\% child-cluster bootstrap intervals. The open diamond is a descriptive pre-cutoff reference, not part of the estimated time trend; exact window counts and slopes appear in Table~\ref{tab:persistence} and the text.}
\label{fig:persistence-full}
\end{figure*}

The primary analysis used 20 children with vocabulary-checklist (CDI) dates defining training cutoffs. Models trained on language collected before each cutoff were evaluated on held-out speech from other people in the same child's environment recorded afterward, using the final checkpoint. There were 67 candidate cutoffs, 65 with comparator sets, and 63 contributing usable future evaluations. The 163 usable future evaluations were grouped by elapsed time (Table~\ref{tab:persistence}): 0--3 months (61 evaluations, 20 children), 3--6 months (48, 17), 6--12 months (35, 10), and more than 12 months (19, 6). Figure~\ref{fig:persistence-full} shows the original longitudinal results and replication. A pre-cutoff reference window from held-out recordings before the training cutoff averaged 0.853 BPW (65 evaluations across 20 children); it was excluded from the analysis of change over time. The primary analysis required at least 2,000 evaluated words per comparison; a 5,000-word threshold was tested separately. Prediction advantage was defined as the average comparison-model loss minus the own-environment model's loss on the same held-out text, in BPW; positive values favor the own-environment model.

\begin{table}[ht]
\centering
\small
\setlength{\tabcolsep}{3pt}
\begin{tabular}{lrrl}
\toprule
Months & Cells & Children & BPW [95\% CI] \\
\midrule
0--3 & 61 & 20 & $0.640\ [0.574,0.705]$ \\
3--6 & 48 & 17 & $0.525\ [0.450,0.600]$ \\
6--12 & 35 & 10 & $0.456\ [0.354,0.556]$ \\
12+ & 19 & 6 & $0.326\ [0.206,0.429]$ \\
\bottomrule
\end{tabular}
\caption{Own-environment prediction advantage for later speech from other speakers in the child's environment. Intervals resample children, not individual evaluation windows.}
\label{tab:persistence}
\end{table}

Fitting within-child slopes across usable future windows gives a decline of $-0.0376$ BPW/month (95\% interval $[-0.0575,-0.0218]$), negative for all 16 children with estimable slopes; four children did not have enough windows to fit a slope. An independent chronological-prefix analysis used 17 children, 31 cutoffs, and 91 cells and gave a slope of $-0.0321$ BPW/month (95\% interval $[-0.0463,-0.0199]$), with 15 of 16 estimable slopes negative. Results were similar with the 5,000-word evaluation threshold ($-0.0365$), validation-selected checkpoints ($-0.0351$), and all-speech scoring ($-0.0360$). Because children also get older as time passes, we cannot tell whether the decline in specialization is due to elapsed time, changes associated with age, or both.

\section{Synthetic Differences and Specialization}\label{app:synthetic-specialization}
We constructed controlled synthetic environments using TinyDialogues with the same generation and preprocessing setup. At each mixing level, eight environments each contributed 100k words; three model seeds per environment gave 192 source-by-target evaluation cases per level (8 training environments $\times$ 8 held-out environments $\times$ 3 seeds). The primary manipulation varied a mixture of discourse and participant composition using $\alpha\in\{0,0.25,0.5,0.75,1\}$. We measured the mean unigram Jensen--Shannon divergence (JSD, bits) between environments rather than assuming that the manipulation produced equal-sized differences.

\begin{table}[ht]
\centering
\small
\setlength{\tabcolsep}{2pt}
\begin{tabular}{lrrrl}
\toprule
Condition & $\alpha$ & JSD & BPW advantage & 95\% CI \\
\midrule
Primary & 0 & .0345 & $-.002$ & $[-.030,.028]$ \\
Primary & .25 & .0426 & $.097$ & $[.010,.193]$ \\
Primary & .50 & .0650 & $.366$ & $[.205,.575]$ \\
Primary & .75 & .1012 & $.826$ & $[.590,1.098]$ \\
Primary & 1.0 & .1524 & $2.107$ & $[1.690,2.465]$ \\
Noun/verb & 1.0 & .0572 & $.289$ & $[.232,.347]$ \\
Noun/category & 1.0 & .1049 & $.857$ & $[.494,1.229]$ \\
Interlocutor & 1.0 & .0542 & $.976$ & $[.584,1.633]$ \\
\bottomrule
\end{tabular}
\caption{Controlled synthetic-environment specialization. The advantage is other-environment minus own-environment content BPW; larger values indicate stronger specialization. JSD is calculated over the top 5,000 unigram types with add-0.5 smoothing. Intervals resample environments.}
\label{tab:synthetic}
\end{table}

\begin{figure*}[t]
\centering
\includegraphics[width=.94\textwidth]{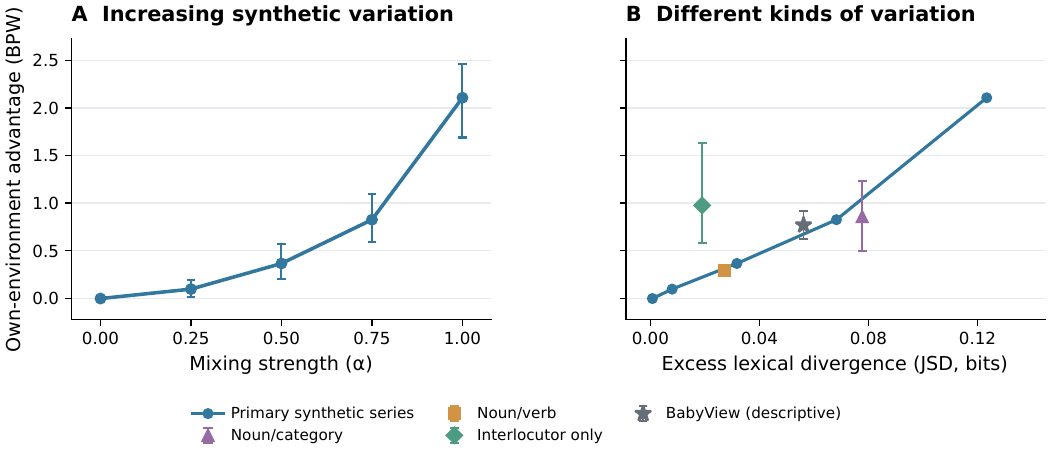}
\caption{Controlled synthetic specialization. \textbf{(A)} Increasing the synthetic mixing parameter $\alpha$ increases the own-environment prediction advantage. \textbf{(B)} The same results plotted against measured excess unigram divergence, alongside three alternative synthetic manipulations and BabyView as a descriptive reference. Excess divergence is the median other-environment minus own-environment JSD. Error bars indicate 95\% environment-cluster bootstrap intervals. The BabyView marker is not part of the controlled synthetic experiment. Different types of changes to the language input can produce different levels of specialization, even when the overall differences between environments are similar.}
\label{fig:synthetic-full}
\end{figure*}

Table~\ref{tab:synthetic} and Figure~\ref{fig:synthetic-full} show the synthetic manipulation and specialization trend. The advantage grows monotonically with the primary manipulation. At similar measured divergence, the original noun/verb and noun/category controls approximately follow the same pattern as the primary mixture. However, the interlocutor-only manipulation yields about 0.976 BPW advantage at JSD 0.0542, approximately 4.1 times the main-series value interpolated at comparable JSD. Hence, measured unigram divergence is not a universal predictor of specialization across different kinds of distributional change. BabyView's 0.769 BPW advantage at mean JSD 0.106 is compatible with the primary synthetic relationship at comparable divergence, although the effect in BabyView is much more concentrated in unseen vocabulary than in these synthetic manipulations. These results show that differences in the composition of language input can produce specialization, but they do not mean that the synthetic environments capture all the properties of real children's language environments.

\section{Raw Scaling for GPT-2 Micro}\label{app:micro-scaling}
\begin{figure*}[t]
\centering
\includegraphics[width=.96\textwidth]{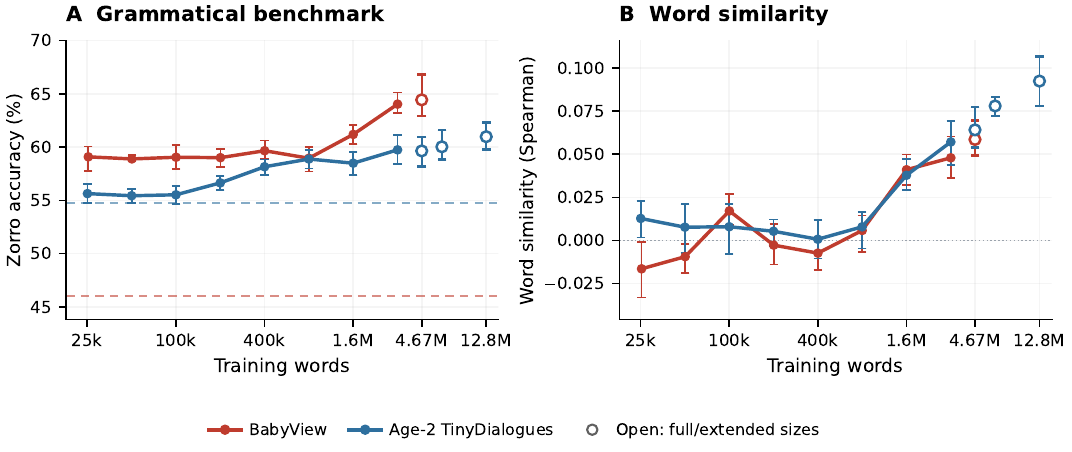}
\caption{Raw benchmark scaling for GPT-2 micro: (A) Zorro accuracy and (B) mean WordSim Spearman correlation over all 7,917 word pairs in five datasets. Lines connect the shared 25k--3.2M training sizes (pointwise 95\% two-stage bootstrap intervals over three corpus draws and three seeds). Hollow markers show the full 4.67M-word BabyView corpus (one draw and a seed-only range), the size-matched age-2 TinyDialogues sample (three draws), and age-2-only extensions at 6.4M and 12.8M words. Color-matched horizontal dashed lines give tokenizer-matched random-initialization Zorro baselines (46.1\% for BabyView and 54.8\% for age-2 TinyDialogues).}
\label{fig:micro-scaling}
\end{figure*}

Figure~\ref{fig:micro-scaling} shows the raw benchmark scaling curves for the 13.9M-parameter GPT-2 micro model, using the same evaluation conventions as the main Figure~\ref{fig:rawscaling}. The full-corpus BabyView anchor has one corpus draw and three seeds (seed-only range), whereas the shared 25k--3.2M series and the size-matched age-2 anchor use three draws and three seeds. The larger age-2-only extensions have no naturalistic counterpart.

\section{Word-Similarity and Low-Data Zorro Checks}\label{app:evaluation-checks}

\subsection{WordSim Evaluation Sensitivity}\label{app:wordsim-sensitivity}
Our primary WordSim evaluation extracts word representations from the final transformer layer, averages representations across subtokens, and computes Spearman correlation with the human judgments on five datasets. The same extraction rule is used for all trained models. Earlier related work such as \citet{feng2026babyscale} instead used the last subtoken and selected the best layer \emph{separately for each test dataset}. The current evaluator also uses a different input prefix and a smaller corpus-specific tokenizer. Scores from the two procedures are hence not directly comparable.

The present evaluations stored correlations at every hidden layer, allowing us to compare layer-selection rules without loading or rescoring models. Table~\ref{tab:wordsim-sensitivity} compares the mean scores for GPT-2 small trained on all 4.67M BabyView words (one corpus draw, three seeds):
\begin{table}[ht]
\centering
\footnotesize
\setlength{\tabcolsep}{3pt}
\begin{tabular}{@{}p{.75\columnwidth}r@{}}
\toprule
Representation/layer convention & Mean $\rho$ \\
\midrule
Final layer (five tests) & 0.050 \\
Two-thirds depth (five tests) & 0.082 \\
MEN-dev-selected (four tests) & 0.044 \\
Best layer per test (five tests) & 0.128 \\
Previous work \citep{feng2026babyscale} (five tests) & 0.153 \\
\bottomrule
\end{tabular}
\caption{WordSim sensitivity to layer selection. The best-layer-per-test-set procedure uses the evaluation targets to choose a layer, so its score is optimistic. The MEN-dev-selected result covers four held-out datasets, unlike the five-dataset means. The v1 result also differs in model, tokenizer, pooling, and input format.}
\label{tab:wordsim-sensitivity}
\end{table}

The older best-layer selection rule also increases the score for an \emph{untrained} BabyView-tokenized small model from $-0.007$ (final layer) to $0.028$. This illustrates why test-set layer selection should not replace the primary fixed-layer evaluation. The archived scores do not separately identify the effect of last-subtoken versus mean-subtoken pooling.

Averaged across training sizes from 800k to 3.2M words,
GPT-2 small models trained on BabyView scored slightly lower
on WordSim than those trained on age-2 TinyDialogues
(a difference of $-0.017$ in mean Spearman correlation).
This difference increased to $-0.057$ when we restricted
the evaluation to word pairs from a shared vocabulary.
Similarly, BabyView scored lower than mixed-age TinyDialogues
by $-0.046$ across all word pairs and $-0.074$ on the
shared-vocabulary subset. These two evaluations use
different sets of word pairs, so their results should be
considered separately. These differences are specific to
the evaluated benchmarks and do not establish that BabyView
is worse for learning word meanings.


\subsection{Random initialization and 25k-word Zorro}\label{app:lowdata-zorro}
Figure~\ref{fig:rawscaling}A shows the random-initialization references for GPT-2 small, matched to each corpus tokenizer and model seed. These anchors are constant across the displayed training sizes because they refer to the same untrained model/tokenizer configurations; they are \emph{not} the nominal 50\% chance line. They differ between corpora, so training gains should be compared with the corresponding initialization rather than with a single universal baseline. Table~\ref{tab:25k-init} summarizes the results at 25k words (three draws and three seeds per corpus).
\begin{table}[ht]
\centering
\small
\setlength{\tabcolsep}{3pt}
\begin{tabular}{@{}lrrr@{}}
\toprule
Corpus / benchmark & Trained & Initial & Gain \\
\midrule
BabyView: Zorro (\%) & 59.44 & 48.41 & +11.03 \\
Age-2 TD: Zorro (\%) & 54.78 & 51.98 & +2.80 \\
BabyView: per-token Zorro (\%) & 60.63 & 47.71 & +12.91 \\
Age-2 TD: per-token Zorro (\%) & 56.59 & 51.22 & +5.37 \\
BabyView: BLiMP (\%) & 54.86 & 49.43 & +5.43 \\
Age-2 TD: BLiMP (\%) & 51.35 & 51.93 & $-0.58$ \\
\bottomrule
\end{tabular}
\caption{25k-word trained versus matched random-initialization performance (GPT-2 small). Gains are in percentage points. The random-initialization reference averages the matched seeds and tokenizer; repeated corpus draws do not create independent initializations.}
\label{tab:25k-init}
\end{table}

The BabyView 25k-word improvement is uneven across phenomena. Selected gains over initialization (percentage points) include subjective pronoun case $+33.5$, negative-polarity-item licensing $+29.5$, local attractors $+26.0$, argument structure $+21.2$, and quantifiers $+14.4$. In contrast, subject--verb agreement ($+0.1$), determiner--noun agreement (approximately zero), and irregular forms ($-0.8$) remain near chance; ellipsis declines ($-13.7$). These are descriptive comparisons, not evidence that broad grammatical rules have been acquired. An existing in-vocabulary Zorro filter gives a similar BabyView result (59.26\% versus 59.44\% unfiltered; age-2: 54.57\% versus 54.78\%). For GPT-2 micro, the corresponding random-initialization Zorro references are 46.1\% (BabyView) and 54.8\% (age-2); these are shown in Figure~\ref{fig:micro-scaling}A.

\section{Age-Composition Comparisons (Supplementary)}\label{app:agecomp}
The age-2 TinyDialogues condition consists of child-directed conversations generated for a two-year-old. The controlled M1 condition combines dialogue targeted at ages 2, 5, and 10 from the same generator family. Its matched comparison with age-2 spans 25k--12.8M words, using three corpus draws and three seeds at each size. 
M1 continues to 51.2M words without a matched age-2 comparison. Because M2 changes both the target ages and the model generating the conversations, we cannot isolate the effect of age. We include its result at $\sim$190M words for context only.

\begin{figure*}[t]
\centering
\includegraphics[width=.96\textwidth]{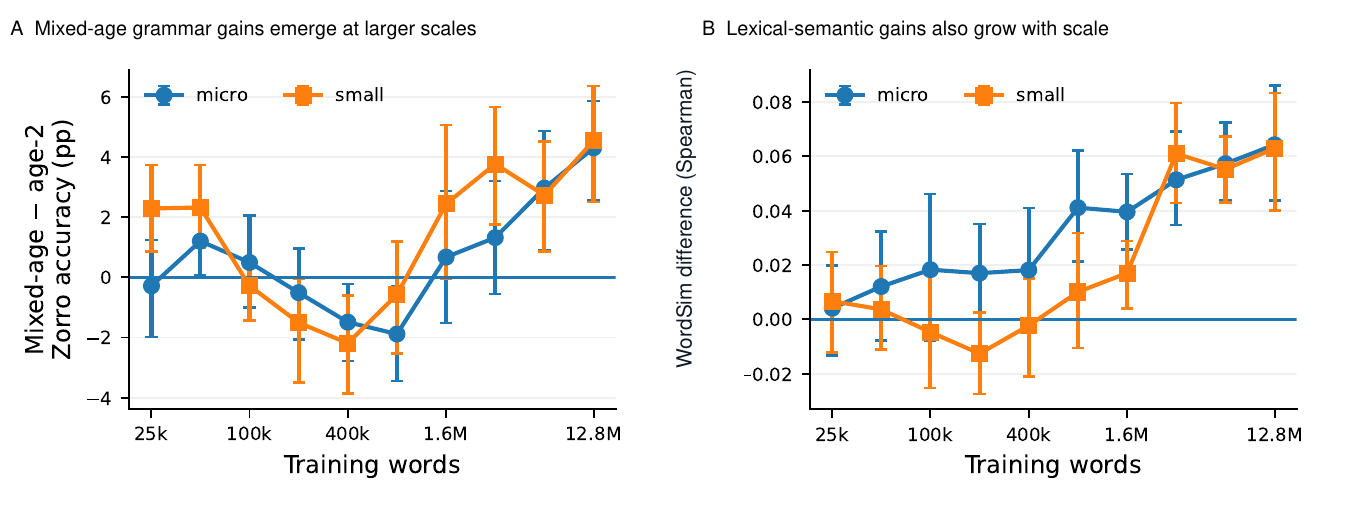}
\caption{Mixed-age minus age-2 TinyDialogues across training sizes: (A) Zorro accuracy differences (percentage points) and (B) word-similarity differences (Spearman). Error bars resample corpus draws and seeds.}
\label{fig:agecomp}
\end{figure*}

Figure~\ref{fig:agecomp} shows the full age-composition comparison. At 25k and 50k words, mixed-age GPT-2 small had Zorro advantages of about $+2.30$ and $+2.33$ percentage points, but at 200k and 400k the differences reversed ($-1.50$ and $-2.19$ points); these apparent low-data deficits weakened in the equal-length item subset. At 3.2M, 6.4M and 12.8M words, the mixed-age small-model advantages were $+3.77$ (95\% interval $[1.77,5.68]$), $+2.74$ ($[0.87,4.52]$), and $+4.55$ ($[2.51,6.38]$) points. 
At larger training sizes, mixed-age data also performed better under the per-token Zorro evaluation and on BLiMP. The smaller GPT-2 micro model showed a similar pattern, although only near the largest training sizes. WordSim combines five benchmark datasets \citep{rubenstein1965rg,finkelstein2002ws353,hill2015simlex,gerz2016simverb,bruni2014men}. Using the same 7,917 word pairs across five benchmarks for both training conditions, mixed-age TinyDialogues scored higher than age-2-only TinyDialogues on WordSim by 0.061 at 3.2M words, 0.055 at 6.4M, and 0.063 at 12.8M (differences in mean Spearman correlation). Because both conditions use synthetic data, these results do not tell us whether mixed-age input is better or worse than naturalistic child-language input.

In held-out text prediction, mixed-age models increasingly
approached the performance of age-2 models on age-2 text
as training data increased. However, age-2-only models
performed increasingly worse than mixed-age models on
held-out mixed-age text. These results suggest that the
relative performance of the two synthetic training
conditions depends on which language distribution
the models are evaluated on.

\section{Fixed-Budget Diversity: Supplementary Tests}\label{app:diversity}
At the primary 100k-word budget, there were 63 training mixtures and 189 runs. The pooled held-out prediction slope was $-0.1236$ BPW per unit $\ln k$ (95\% interval $[-0.1356,-0.1116]$; Monte-Carlo permutation $p=0.0001$), whereas the primary Zorro slope was $+0.2003$ percentage points per $\ln k$ (HC1 95\% interval $[-0.282,+0.682]$; $p=0.2788$). In leave-one-dataset-out analyses, $R^2$ was 0.673 using the number of contributing children ($\ln k$), 0.930 using unigram JSD, 0.719 using vocabulary coverage, and 0.974 using five distributional descriptors together. These results show relationships between training-data properties and prediction performance, but do not establish cause and effect or show whether the findings generalize to new children.

At the secondary 250k-word budget (27 mixture datasets and 81 runs), pooled held-out BPW improved with additional children (slope $-0.1465$; $p=0.0001$). The primary Zorro slope was $+0.594$ points (95\% interval $[-0.077,1.265]$; $p=0.0854$); per-token-normalized Zorro was positive ($p=0.044$), while BLiMP showed no clear effect ($p\approx0.37$). These supplementary tests expand the Section~\ref{diversity-improves-pooled-distribution-prediction-more-clearly-than-grammatical-benchmark-performance} analysis. Hence, the primary 100k-word conclusion should not be generalized to a claim that increasing diversity never helps grammaticality.

\section{Child-Speech Removal Control}\label{app:childspeech}
As a control for Section~\ref{specialization-extends-beyond-the-target-childs-speech},
we compared two training conditions for 10 children with
enough recorded speech: removing the target child's own
utterances, or removing the same amount of speech at random.
Each condition used three model seeds, and results were
compared within children and seeds. On pooled held-out BabyView text, removing the child's
speech resulted in 9.761 BPW, compared with 9.504 BPW
under random removal (difference $+0.257$ BPW;
10/10 children; exact sign-flip $p=0.002$).
On held-out text from the child's own environment,
the corresponding values were 8.845 and 8.562 BPW
(difference $+0.283$ BPW; $p=0.002$). In contrast, Zorro accuracy was similar between
conditions (57.89\% versus 58.02\%;
difference $-0.12$ percentage points, $p=0.77$),
as was BLiMP (52.02\% versus 52.25\%;
difference $-0.24$ points, $p=0.28$).
Thus, the child's own speech contributes useful
information for predicting held-out language, but
we find no corresponding improvement on these
grammatical benchmarks.

\section{Developmental Outcomes (Supplementary)}\label{app:developmental}

Following earlier work on predicting when children learn words \citep{chang2022word,portelance2023aoa}, we tested whether the order in which language models learn words helps predict children's age of word acquisition in Wordbank
\citep{frank2017wordbank}. Our baseline already included word frequency, concreteness, and word class. Adding the model's learning order did not improve predictions according to our predefined criterion ($\Delta\mathrm{AIC}=+0.66$, reported interval $[-2.91,+1.51]$; improvement required a value below $-2$). The additional predictor also had only a small correlation with children's age of acquisition after accounting for the baseline factors ($r=-0.085$, 95\% interval $[-0.154,-0.015]$).

In a separate analysis using child-specific vocabulary reports (CDI; \citealp{fenson2007cdi}) from 20 children (65 reports), models trained on a child's own language environment showed a small advantage in distinguishing words the child could produce from words they could not ($\Delta\mathrm{AUC}=0.0229$, permutation $p=0.0006$). However, a predictor based only on word frequency performed slightly better ($\Delta\mathrm{AUC}=0.0298$), and the model's advantage disappeared and reversed after accounting for frequency ($\Delta\mathrm{AUC}=-0.0147$). These results do not provide convincing evidence that our models capture how children actually learn vocabulary.

\section{Computational Resources and Implementation Details}\label{app:compute}

\paragraph{Hardware and compute.}
Model training and most GPU evaluations ran on eight single-node RunPod instances with 1--8 NVIDIA H100 80\,GB or 2 NVIDIA H200 GPUs (CUDA~12.8). A local workstation with an RTX~3090~Ti 24\,GB handled two additional GPU evaluation jobs. Corpus preparation, tokenizer training, KN-5 baselines, and statistical analyses used workstation or pod CPUs. The experiment ledger accounts for approximately 429.5 GPU-hours of training and GPU-based evaluation: 411.8 hours in experiment families contributing to this paper and 17.7 hours in pilots or other investigations (Table~\ref{tab:compute}). GPU time was estimated from 10-second \texttt{nvidia-smi} activity samples where available; for some jobs we instead sum process wall-clock times, which can overcount overlapping GPU work. Thus, these are approximate, not uniformly measured, GPU-active hours. The eight GPU pods were allocated for approximately 807 GPU-hours from creation to termination (one creation time is estimated), including setup, CPU-only work, and idle periods. CPU-only compute was not separately metered.

\begin{table}[t]
\centering\small
\begin{tabularx}{\columnwidth}{@{}>{\RaggedRight\arraybackslash}Xrr@{}}
\toprule
Experiment family & Runs & GPU-h \\
\midrule
Main scaling and closeout checks & 390 & 35.4 \\
Mixed-age TinyDialogues scaling & 636 & 239.6 \\
Large-data replication (plus 24 replays) & 24 & 14.0 \\
Recipe sensitivity grid & 600 & 20.4 \\
Individual environments and prefixes & 525 & 20.9 \\
Diversity, reference data, and AoA & 494 & 23.8 \\
Robustness experiments & 534 & 18.3 \\
Pre-CDI child models & 201 & 20.9 \\
Controlled synthetic specialization & 244 & 11.0 \\
GPU re-scoring and persistence (no new models) & -- & 7.3 \\
Pilots and other investigations & $\approx$1{,}500 & 17.7 \\
\midrule
Total & $\approx$5{,}100 & $\approx$429.5 \\
\bottomrule
\end{tabularx}
\caption{Approximate GPU-work accounting. GPU-hours combine sampled utilization and process-time proxies; training-run counts exclude the separate replay jobs. Evaluation time is included. Allocated RunPod time was approximately 807 GPU-hours.}
\label{tab:compute}
\end{table}

\paragraph{Models and optimization.}
We use Python~3.11, PyTorch~2.11.0+cu128, \texttt{transformers}~5.12.1, \texttt{tokenizers}~0.22.2, and NumPy~2.4.6. GPT-2 micro has 6 layers, 8 attention heads, and width 384; GPT-2 small has 12 layers, 12 heads, and width 768. All use 1{,}024-token contexts and tied embeddings; parameter counts vary with vocabulary size as reported in \S\ref{models-and-training}. AdamW uses $(\beta_1,\beta_2)=(0.9,0.95)$, $\epsilon=10^{-8}$, and weight decay on matrix-valued parameters only. Each optimizer step uses 16{,}384 tokens (16 blocks of 1{,}024), without gradient accumulation, with gradient clipping at norm~1. The learning rate warms up linearly over 3\% of steps and then decays by a cosine schedule to 10\% of its peak. Training uses bf16 autocast and fp32 master weights. The prepared block order is fixed across passes, without reshuffling; model seeds are 0--2. The held-out selection split is scored every 5\% of training steps. Fixed-recipe and tuned hyperparameters are reported in Appendix~\ref{app:recipe}.

\paragraph{Tokenizers and baseline.}
Tokenizers use byte-level BPE (GPT-2 pre-tokenization; minimum frequency~2). Speaker tags and utterance separators are added as special tokens. The shared 16k-type tokenizer is used for individual-child analyses; the 7.5k-type scaling tokenizers are trained on corpus-specific reserved splits excluded from model training and final evaluation. KN-5 uses our own modified interpolated Kneser--Ney implementation \citep{kneser1995,chen1999}, with three count-based discounts per order and a uniform base reserving mass for \texttt{<unk>}. Text is lowercased and whitespace-tokenized; n-gram contexts do not cross utterance boundaries.

\paragraph{Evaluation.}
Content bits per word are obtained by summing negative log-probabilities over content tokens in non-overlapping 1{,}024-token windows, dividing by $\ln 2$ and by the word count; format tokens are excluded. Zorro and BLiMP minimal pairs are scored by summed full-sentence log-probability, with ties counting as half credit; BLiMP uses the BabyLM-2026 vocabulary-filtered release. Word similarity uses final-layer hidden states averaged across subtokens, cosine similarity, and Spearman correlation. Appendix~\ref{app:stats} gives resampling methods; analysis scripts record seeds and repository commits. We use transcripts from the BabyView~2026.1 release with our own quality, language, and speaker filters, without rerunning the upstream transcription pipeline. TinyDialogues data are reused from prior work \citep{feng2024tinydialogues}.

\end{document}